\documentclass[sigconf,natbib=false]{acmart}

\usepackage{booktabs} 
\usepackage{fancyhdr}
\usepackage{enumitem}
\usepackage{subcaption}
\usepackage{multirow}
\usepackage{listings}
\usepackage{xcolor}
\usepackage{amsmath}

\usepackage{mathrsfs,amssymb}
\usepackage[boxed,ruled]{algorithm2e}
\SetKwRepeat{Do}{do}{while}
\usepackage{makecell}
\usepackage[para,online,flushleft]{threeparttable}
\usepackage{mathtools}
\usepackage{bm} 
\DeclarePairedDelimiter{\ceil}{\lceil}{\rceil}
\DeclareMathAlphabet{\mathcal}{OMS}{cmsy}{m}{n}

\RequirePackage[abbreviate=true, dateabbrev=true, isbn=true, doi=true, urldate=comp, url=true, maxbibnames=9, backref=false, backend=biber, style=ACM-Reference-Format, language=american]{biblatex}

\acmYear{2018}
\copyrightyear{2018}
\setcopyright{acmcopyright}
\acmConference{FPGA'18}{Feb. 25--27, 2018}{Monterey, CA, USA}\acmPrice{15.00}
\acmBooktitle{FPGA'18: 2018 ACM/SIGDA International Symposium on Field-Programmable Gate Arrays, February 25--27, 2018, Monterey, CA, USA}
\acmPrice{15.00}
\acmDOI{10.1145/3174243.3174253}
\acmISBN{978-1-4503-5614-5/18/02}

\begin{document}

\title{C-LSTM: Enabling Efficient LSTM using Structured Compression Techniques on FPGAs}

\author{Shuo Wang$^{1,+}$, Zhe Li$^{2,+}$, Caiwen Ding$^{2,+}$, {Bo Yuan}$^3$, Qinru Qiu$^2$, Yanzhi Wang$^2$ and Yun Liang$^{1,}$}
\authornote{Corresponding author.}
\affiliation{$^+$These authors contributed equally}
\affiliation{$^1$Center for Energy-Efficient Computing \& Applications (CECA), School of EECS, Peking University, China\\ $^2$Dept. of Electrical Engineering \& Computer Science, Syracuse University, Syracuse, NY, USA\\ {$^3$Dept. of Electrical Engineering, City University of New York, NY, USA\\$^1$\{shvowang,ericlyun\}@pku.edu.cn, $^2$\{zli89,cading,qiqiu,ywang393\}@syr.edu}, $^3$byuan@ccny.cuny.edu}

\renewcommand{\shortauthors}{Shuo Wang, Zhe Li, Caiwen Ding et al.}

\begin{abstract}
Recently, significant accuracy improvement has been achieved for acoustic recognition systems by increasing the model size of Long Short-Term Memory (LSTM) networks. Unfortunately, the ever-increasing size of LSTM model leads to inefficient designs on FPGAs due to the limited on-chip resources. The previous work proposes to use a pruning based compression technique to reduce the model size and thus speedups the inference on FPGAs. However, the random nature of the pruning technique transforms the dense matrices of the model to highly unstructured sparse ones, which leads to unbalanced computation and irregular memory accesses and thus hurts the overall performance and energy efficiency. 

In contrast, we propose to use a structured compression technique which could not only reduce the LSTM model size but also eliminate the irregularities of computation and memory accesses. This approach employs block-circulant instead of sparse matrices to compress weight matrices and reduces the storage requirement from $\mathcal{O}(k^2)$ to $\mathcal{O}(k)$. Fast Fourier Transform algorithm is utilized to further accelerate the inference by reducing the computational complexity from $\mathcal{O}(k^2)$ to $\mathcal{O}(k\text{log}k)$. The datapath and activation functions are quantized as 16-bit to improve the resource utilization. More importantly, we propose a comprehensive framework called C-LSTM to automatically optimize and implement a wide range of LSTM variants on FPGAs. According to the experimental results, C-LSTM achieves up to 18.8X and 33.5X gains for performance and energy efficiency compared with the state-of-the-art LSTM implementation under the same experimental setup, and the accuracy degradation is very small.
\end{abstract}


\maketitle

\section{Introduction}

Recurrent neural networks (RNNs) represent an important class of neural networks that contain cycles to carry information across neurons while reading inputs. Long Short-Term Memory (LSTM), one of the most popular types of RNNs, achieves great success in the domains such as speech recognition, machine translation, scene analysis, etc.~\cite{sak2014long}. However, the significant recognition accuracy improvement comes at the cost of increased computational complexity of larger model size~\cite{gers2000recurrent}. Therefore, customized hardware acceleration is increasingly important for LSTMs, as exemplified by recent works on employing GPUs~\cite{cui2012accurate,liang2015efficient}, FPGAs~\cite{han2017ese,li2015fpga} and ASICs~\cite{esser2016convolutional} as accelerators to speedup LSTMs. 

Among the numerous platforms, FPGA has emerged as a promising solution for hardware acceleration as it provides customized hardware performance with flexible reconfigurability. By creating dedicated pipelines, parallel processing units, customized bit width, and etc., application designers can accelerate many workloads by orders of magnitude using FPGAs~\cite{rupnow2011high}. More importantly, High-level Synthesis (HLS) has greatly lowered the programming hurdle of FPGAs and improved the productivity by raising the programming abstraction from tedious RTL to high-level languages such as C/C++~\cite{HLS:2011:Cong} and OpenCL~\cite{Wang:2017:FlexCL}.

While the benefits of FPGAs is clear, it is still challenging to design efficient designs for LSTMs on FPGAs mainly for two reasons. On one hand, the capacity of the FPGA on-chip memory (a few or tens of Mb on-chip memory) is usually not large enough to store all the weight matrices of a standard LSTM inference model (e.g. hundreds of Mb). Although the previous work ESE~\cite{han2017ese} proposes to use the parameter pruning based compression technique to compress the dense weight matrices in the LSTM model into sparse ones, the sparse matrices need extra storage and processing units to store and decode the indices of the non-zero data, respectively. The skewed distribution of the data is likely to cause unbalanced workloads among parallel compute units. Therefore, the benefits of unstructured model compression is diminished by the sparsity of weight matrices. On the other hand, the computational complexity among the operators of the LSTMs is highly skewed and the data dependencies between operator are complicated. So, it is difficult to evenly allocate computing resources under the FPGA resource constraints while guaranteeing the complex data dependencies.

In this work, we propose to compress the weight matrices in the LSTM inference model in a structured manner by using block-circulant matrix~\cite{pan2012structured}. The circulant matrix is a square matrix, of which each row (column) vector is the circulant reformat of the row (column) vector. Any matrix could be transformed into a set of circulant submatrices a.k.a. block-circulant matrices. Therefore, by representing each block-circulant matrix with a vector, the storage requirement could be reduced from $\mathcal{O}(k^2)$ to $\mathcal{O}(k)$ if the block (vector) size is $k$. Since the compressed weight matrices are still dense, the block-circulant matrix based compression is amenable to hardware acceleration on FPGAs. In order to further speed up the computation of LSTMs, we propose to accelerate the most computation-intensive circulant convolution operator by applying Fast Fourier Transform (FFT) algorithm to reduce the computational complexity from  $\mathcal{O}(k^2)$ to $\mathcal{O}(k\text{log}k)$. 

After the model is compressed, we propose an automatic optimization and synthesis framework called C-LSTM to port efficient LSTM designs onto FPGAs. The framework is composed of model training and implementation flows. The former one is in charge of iteratively training the compressed LSTM model and exploring the trade-offs between compression ratio and prediction accuracy. As for the model implementation, it mainly consists of two parts which are (1) template generation and (2) automatic LSTM synthesis framework. For the former part, after analyzing a wide range of LSTM algorithms, we generalize a suite of LSTM primitive operators which is general enough to accommodate even the most complicated LSTM variant~\cite{sak2014long}. Then, a suite of highly optimized C/C++ templates of the primitive operators are manually generated by walking through a series of optimizations such as datapath and activation quantization, DFT-IDFT decoupling and etc. As for the latter part, the well-trained LSTM inference model is first analyzed and transformed into a directed acyclic dependency graph, where each node represents an operator and each edge indicates the associated data dependency between two operators. Secondly, we propose a specialized pipeline optimization algorithm considering both coarse-grained and fine-grained pipelining schemes to schedule the operators into appropriate stages. In the third step, we use an accurate performance and resource model to enable a fast design space exploration for optimal design parameters. Lastly, the scheduling results and optimization parameters are fed to code generator and backend toolchain as to implement the optimized LSTM accelerator design on FPGAs.

Overall, the contributions of this paper are listed as:
\begin{itemize}
\item{} We employ the block-circulant matrices based structured compression technique for LSTMs which largely reduces the computation complexity and memory footprint without incurring any computation and memory access irregularities. This method results in both compression and acceleration of the LSTM models.

\item{} We develop a general LSTM optimization and synthesis framework C-LSTM to enable automatic and efficient implementations of a wide range of LSTM variants on FPGAs. The framework mainly consists of a suite of highly optimized C/C++ based templates of primitive operators and an automatic LSTM synthesis flow.

\item{} We present efficient implementations of LSTMs which achieve up to 18.8X and 33.5X gains in performance and energy efficiency, respectively, compared with the state-of-the-art. The proposed implementations incur very small accuracy degradation.

\end{itemize}

 \begin{figure}
  \centering
  \includegraphics[width=0.9\columnwidth]{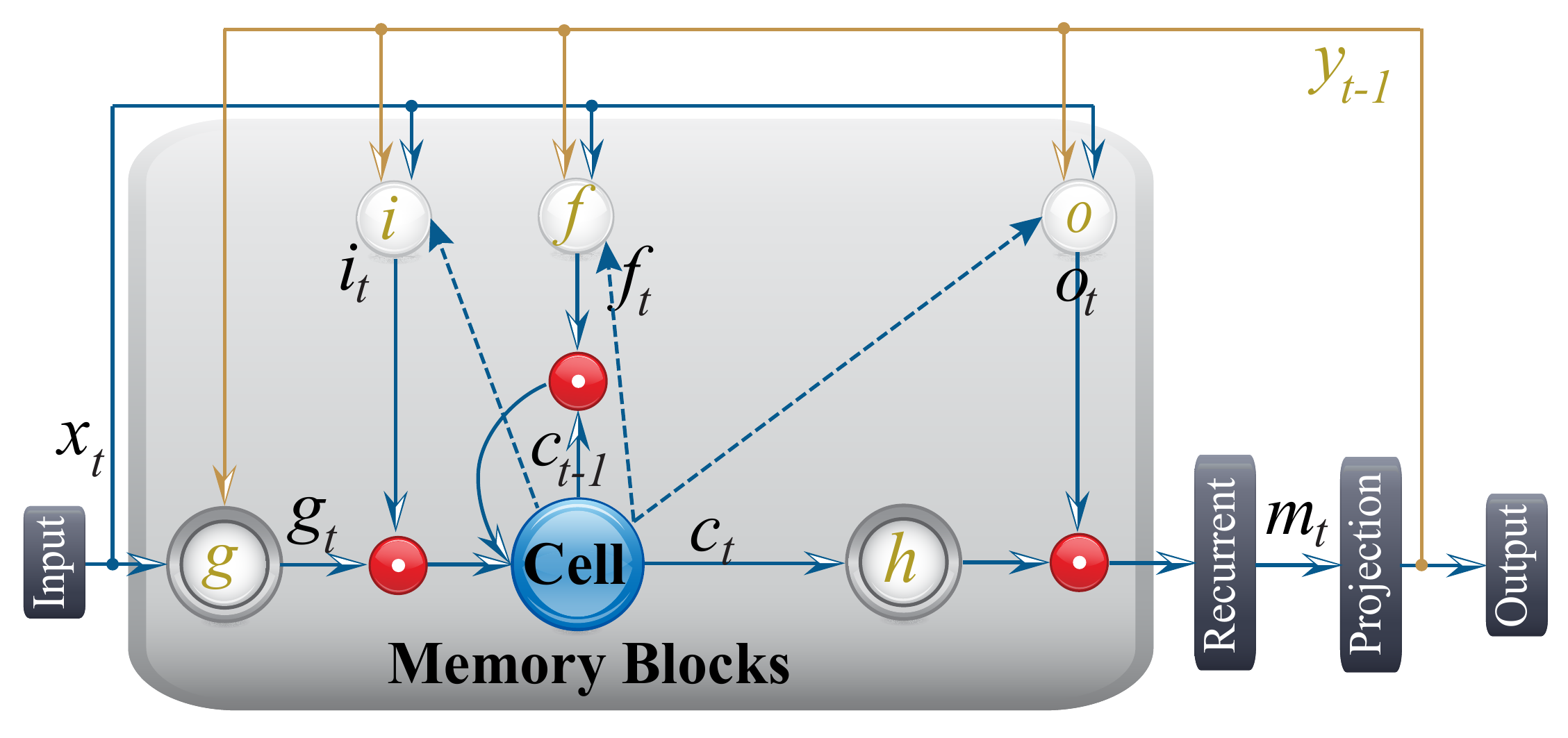}  
  \caption{An LSTM based RNN architecture.}
  \label{fig:LSTM}
 \end{figure}

\section{LSTM Background}
 
LSTM is a key component of the acoustic model in modern large-scale automatic speech recognition (ASR) systems~\cite{gers2000recurrent,sak2014long}, and also the most computation and memory-intensive part. Due to the complicated and flexible data dependencies among gates, cells, and outputs, a lot of LSTM variants have been proposed. In this paper, we use a widely deployed variant called Google LSTM~\cite{sak2014long} as an example throughout this paper without loss of generality. The architecture details of the Google LSTM is shown in Figure~\ref{fig:LSTM}. The LSTM accepts an input sequence $\mathbb{X}= (\mathbf{x}_1; \mathbf{x}_2; \mathbf{x}_3; ...; \mathbf{x}_T)$ (each of $\mathbf{x}_t$ is a vector corresponding to time $t$) with the output sequence from last step $\mathbb{Y}^{T-1} = (\mathbf{y}_0; \mathbf{y}_1; \mathbf{y}_2; ...; \mathbf{y}_{T-1})$ (each of $\mathbf{y}_t$ is a vector). The input of Google LSTM at time $t$ depends on the output at $t-1$. The LSTM contains a special memory cell storing the temporal state of the network. It also contains three special multiplicative units which are input, output and forget gates. The output sequence $\mathbb{Y} = (\mathbf{y}_1;\mathbf{y}_2; \mathbf{y}_3; ...; \mathbf{y}_T )$ is computed by using the following equations iteratively from $t = 1$ to $T$:
\begin{subequations}\label{eqn:model}
\begin{align}
    \mathbf{i}_t &= \sigma(\mathbf{W}_{ix}\mathbf{x}_t +\mathbf{W}_{ir}\mathbf{y}_{t-1} + \mathbf{W}_{ic}\mathbf{c}_{t-1}+\mathbf{b}_i), \\
    \mathbf{f}_t &= \sigma(\mathbf{W}_{fx}\mathbf{x}_t +\mathbf{W}_{fr}\mathbf{y}_{t-1} + \mathbf{W}_{fc}\mathbf{c}_{t-1}+\mathbf{b}_f), \\
    \mathbf{g}_t &= \sigma(\mathbf{W}_{cx}\mathbf{x}_t + \mathbf{W}_{cr}\mathbf{y}_{t-1} + \mathbf{b}_c), \\
    \mathbf{c}_t &= \mathbf{f}_t \odot \mathbf{c}_{t-1} + \mathbf{g}_t \odot \mathbf{i}_t, \\
    \mathbf{o}_t &= \sigma(\mathbf{W}_{ox}\mathbf{x}_t + \mathbf{W}_{or}\mathbf{y}_{t-1} + \mathbf{W}_{oc}\mathbf{c}_{t}+\mathbf{b}_o), \\
    \mathbf{m}_t &= \mathbf{o}_t \odot \mathbf h(\mathbf{c}_t), \\
    \mathbf{y}_t &= \mathbf{W}_{ym}\mathbf{m}_t,
\end{align}
\end{subequations}
where symbols $\mathbf{i}$, $\mathbf{f}$, $\mathbf{o}$, $\mathbf{c}$, $\mathbf{m}$, and $\mathbf{y}$ are respectively the input gate, forget gate, output gate, cell state, cell output, and a projected output;
the $\odot$ operator denotes the element-wise multiplication, and the $+$ operator denotes the element-wise addition.
The $\mathbf{W}$ terms denote weight matrices (e.g. $\mathbf{W}_{ix}$ is the matrix of weights from the input vector $\mathbf{x}_t$ to the input gate), and the $\mathbf{b}$ terms denote bias vectors. 
Please note $\mathbf{W}_{ic}$, $\mathbf{W}_{fc}$, and $\mathbf{W}_{oc}$ are diagonal matrices for peephole connections, thus they are essentially a vector, and the matrix-vector multiplication like $\mathbf{W}_{ic}\mathbf{c}_{t-1}$ can be calculated by the $\odot$ operator.
$\sigma$ is the logistic activation function and $h$ is a user-defined activation function. Here we use hyperbolic tangent (tanh) activation function as $h$. Overall, we have nine matrix-vector multiplications (excluding peephole connections which can be calculated by $\odot$). In one gate/cell, $\mathbf{W}_{\ast x}\mathbf{x}_t + \mathbf{W}_{\ast r}\mathbf{y}_{t-1}$ can be combined/fused in one matrix-vector multiplication by concatenating the matrix and vector as $\mathbf{W}_{\ast (xr)}[{\mathbf{x}_{t}, \mathbf{y}_{t-1}}]$. 





\section{Structured Compression}
Deep neural networks (DNNs) bear a significant amount of redundancy~\cite{han2015deep} and thus model compression is a natural method to mitigate the computation and memory storage requirements for the hardware implementations on FPGAs. In this section, we propose to employ a structured compression technique to compress the weight matrices of LSTM model by using block-circulant matrices. We first introduce the block-circulant matrix and then integrate it with the inference and training algorithms of LSTMs. In the last, we explore the trade-offs between compression ratio and prediction error rate.

\subsection{Block-Circulant Matrix}
\begin{figure}[t]
    \centering
    \includegraphics[width=\columnwidth]{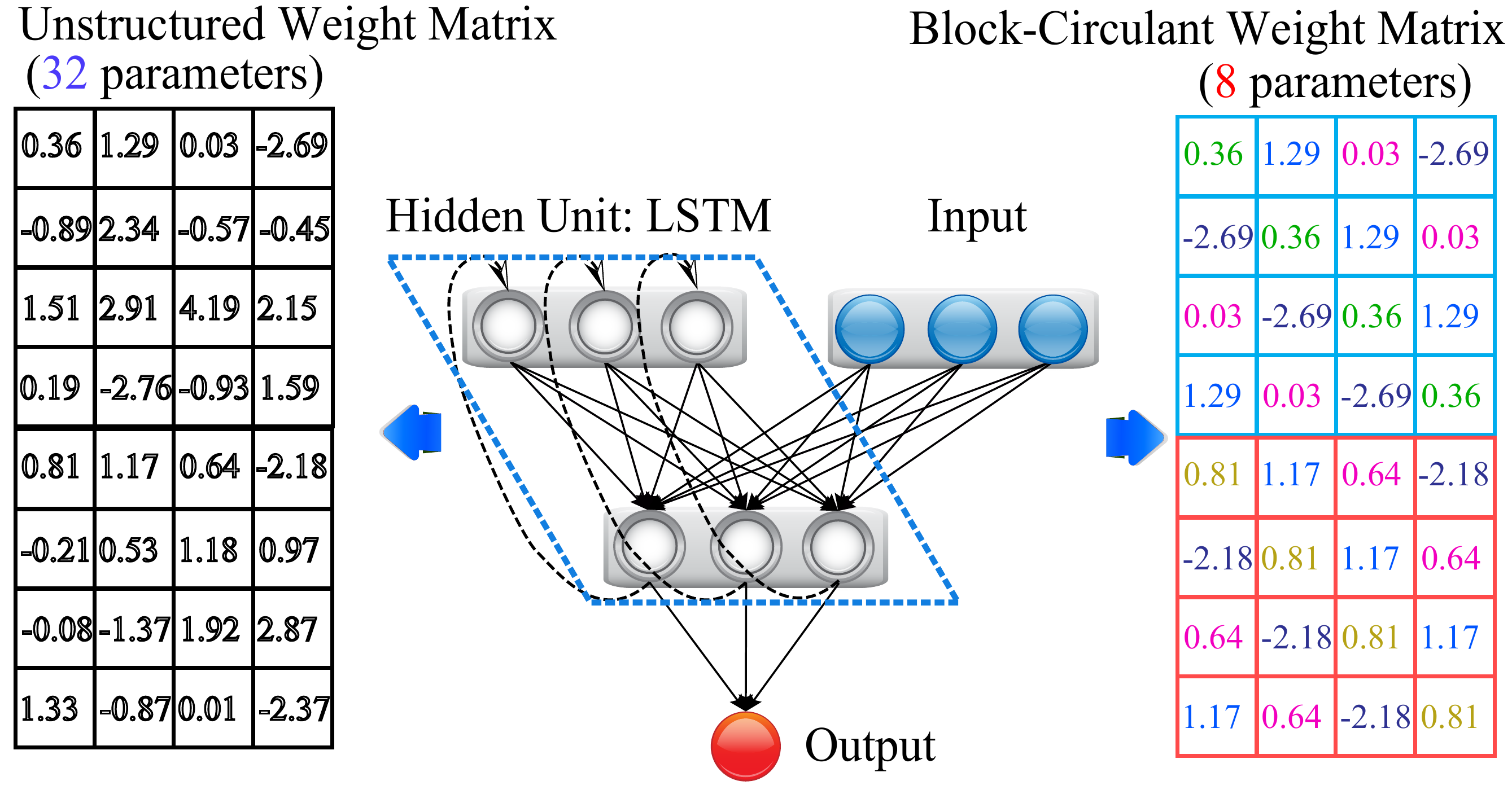} 
    \caption{Block-circulant matrices for weight representation.}
    \label{fig:Intro_Block_Matrix}
 \end{figure}

 The circulant matrix is a square matrix whose each row (or column) vector is the circulant reformat of the row (or column) vectors~\cite{cheng2015exploration,pan2012structured}. Any matrix could be transformed into a set of circulant submatrices (blocks) and we define the transformed matrix as a block-circulant matrix. For example, Figure~\ref{fig:Intro_Block_Matrix} shows that the $8\times4$ weight matrix (on the left) is reformatted into a block-circulant matrix containing two $4\times4$ circulant matrices (on the right). Since each row vector of the circulant submatrix is a reformat of the first row vector, we could use a row vector to represent a circulant submatrix. Therefore, the first obvious benefit of the block-circulant matrix is that the number of parameters in each weight matrix is reduced by a factor of the block size $\mathcal{O}(k)$. As for the example in Figure~\ref{fig:Intro_Block_Matrix}, the $8\times4$ weight matrix (on the left) holding 32 parameters is reduced to two $4\times4$ circulant matrices (on the right) containing only 8 parameters, which easily leads to 4X model size reduction. 

Intuitively, the model compression ratio is determined by the block size of the circulant submatrices: larger block size leads to higher compression ratio and vice versa. However, high compression ratio may degrade the prediction accuracy. Specifically, a larger block size should be selected to achieve a higher compression ratio but lower accuracy and the smaller block size provides higher accuracy but less compression ratio. The block size is 1 if no compression is utilized. It is necessary to note that block-circulant matrix based DNNs have been proved to asymptotically approach the original networks in accuracy with mathematical rigor~\cite{zhao2017theoretical}. Therefore, if the compression ratio is selected properly, the accuracy loss would be negligible. The trade-offs between compression ratio and predication accuracy are discussed in Section~\ref{sec:trade-off}

\subsection{Inference and Training Algorithms}\label{sec:inference_and_training}


The primary idea of introducing block-circulant matrix into LSTM model is to partition a $m \times n$ weight matrix $\bf{W}$ into $p \times q$ blocks, where $p=\frac{m}{k}$, $q=\frac{n}{k}$ and each block is a $k \times k$ circulant matrix. With bias and activation function omitted, the forward propagation process of LSTM model in the inference phase is then given by:
\begin{equation} \label{eqn:weightmult}
\mathbf{a}
=
\mathbf{Wx} 
\iff
\begin{bmatrix}
         \sum_{j=1}^q \mathbf{W}_{1j} \mathbf{x}_j   \\
         \sum_{j=1}^q \mathbf{W}_{2j} \mathbf{x}_j   \\
         \dots \\
         \sum_{j=1}^q \mathbf{W}_{pj} \mathbf{x}_j  
\end{bmatrix}
=
\begin{bmatrix}
         \mathbf{a}_1   \\
         \mathbf{a}_2   \\
         \dots \\
         \mathbf{a}_p
\end{bmatrix},
\end{equation}
where $\mathbf{a}_i$ is a column vector. Since each circulant matrix $\mathbf{W}_{ij}$ could be simplified as a vector $\mathbf{w}_{ij}$, i.e., $\mathbf{w}_{ij}$ is the first row vector of $\mathbf{W}_{ij}$, the structure of block-circulant matrix enables the use of Fast Fourier Transform (FFT) algorithm to speed up the circulant convolution $\sum_{j=1}^q \mathbf{W}_{ij} \mathbf{x}_j$. Therefore the Equation~(\ref{eqn:weightmult}) can be performed as:
\begin{equation}
\label{eqn:core}
 \mathbf{a}_i=\sum_{j=1}^q \mathcal{F}^{-1}[\mathcal{F}(\mathbf{w}_{ij})~\odot~ \mathcal{F}(\mathbf{x}_{j})],
\end{equation}
where $\mathcal{F(\cdot)}$ is the Discrete Fourier Transform (DFT) operator, $\mathcal{F}^{-1}(\cdot)$ is the inverse DFT (IDFT) operator, and $\odot$ is the element-wise multiply operator. Therefore, after applying FFT algorithm to the circulant convolution, the computational complexity of the LSTM inference model is reduced from $\mathcal{O}(pqk^2)$ to $\mathcal{O}(pqk\log k)$, meaning that the computational complexity of the LSTM inference model is reduced by a factor of $\mathcal{O}(\frac{k}{\text{log}k})$.


The backward propagation process in the training phase can also be implemented using block-circulant matrices. Here we use $a_{il}$ to denote the $l$-th output element in $\mathbf{a}_i$, and $L$ to represent the loss function. Then by using the chain rule we can derive the backward propagation process as follows:
\begin{gather}
\frac{\partial L}{\partial \mathbf{w}_{ij}} 
= \sum_{l=1}^k \frac{\partial L}{\partial a_{il}} \frac{\partial a_{il}}{\partial \mathbf{w}_{ij}}
= \frac{\partial L}{\partial \mathbf{a}_i} \frac{\partial \mathbf{a}_i}{\partial \mathbf{w}_{ij}},\\
\frac{\partial L}{\partial \mathbf{x}_{j}} 
= \sum_{i=1}^p\sum_{l=1}^k \frac{\partial L}{\partial a_{il}} \frac{\partial a_{il}}{\partial \mathbf{x}_{j}}
= \sum_{i=1}^p \frac{\partial L}{\partial \mathbf{a}_i} \frac{\partial \mathbf{a}_i}{\partial \mathbf{x}_j}.
\end{gather}
 where $\frac{\partial \mathbf{a}_i}{\partial \mathbf{w}_{ij}}$ and $\frac{\partial\mathbf{a}_i}{\partial \mathbf{x}_j}$ are proved to be block-circulant matrices~\cite{zhao2017theoretical}. Thus, $\frac{\partial L}{\partial \mathbf{w}_{ij}}$ and $\frac{\partial L}{\partial \mathbf{a}_i} \frac{\partial \mathbf{a}_i}{\partial \mathbf{x}_j}$ can be calculated similarly as Equation (\ref{eqn:core})
 with the same computational complexity. The details of the training procedure for a fully-connected layer in DNNs are presented in~\cite{ding2017circnn,wang2018towards} and also applicable to the LSTM based RNNs. 

\subsection{Compression and Accuracy Trade-offs}\label{sec:trade-off}
The block-circulant matrix based LSTM inference model enables a comprehensive tuning of model compression ratio by varying the block size $k$, thus leading to fine-grained trade-offs among the model size, computational complexity, and prediction accuracy. The proposed inference model of Google LSTM~\cite{sak2014long} is evaluated on the widely used TIMIT dataset \cite{garofolo1993darpa}. Similar to \cite{graves2013hybrid}, the audio data of TIMIT is preprocessed using a Fourier transform based filterbank with 50 coefficients (plus energy) distributed on a mel-scale, together with their first and second temporal derivatives. The number of features of the input speech and the architecture of Google LSTM used in this work is the same as ESE~\cite{han2017ese}. It is necessary to note that we use the widely adopted Phone Error Rate (PER) as the metric for the model prediction accuracy. The lower the PER value is, the higher the model prediction accuracy is and vice versa.

Table~\ref{tbl:performance} presents the details of the trade-offs among three different metrics of Google LSTM using the block-circulant matrix based structured compression technique. We observe that the number of model parameters decreases linearly as the block size increases. Meanwhile, the PERs of different models do not have a severe degradation. For the block-circulant matrix based LSTM with block size of 2, the PER is even lower than non-compressed LSTM model whose block size is 1. 
For the LSTM models with block size of 8 and 16, we achieve 7.6X and 14.6X model size reduction and the computational complexity is reduced by factors of 2.6X and 3.7X while the PERs are only $0.32\%$ and $1.23\%$ higher than the non-compressed one, respectively. Therefore, we choose the compressed models of Google LSTM with block sizes of 8 and 16 to be further studied in this work. 




\begin{table}[!t]
   \centering
   \caption{Comparison among different LSTM models.}\label{tbl:performance}
       \begin{tabular}{|c|c|c|c|c}
           \hline
           \makecell{Block\\Size} & \makecell{\#Model \\Parameters} & \makecell{Computational \\ Complexity} & \makecell{PER\ /\ PER\\Degradation (\%)}\\ \hline
           1  & $8.01M$ & 1    & $24.15$ / $\ \ 0.00$ 
           \\ \hline
           2  & $4.03M$ & 0.50 & $24.09$ / $-0.06$\\ \hline
           4  & $2.04M$ & 0.50 & $24.23$ / $\ \ 0.08$\\ \hline
           8  & $1.05M$ & 0.39 & $24.57$ / $\ \ 0.32$\\ \hline
           16 & $0.55M$ & 0.27 & $25.48$ / $\ \ 1.23$\\ \hline
            
       \end{tabular}
\end{table}


\section{FPGA Acceleration} \label{sec:hardware_acc}
In this section, we start by introducing a set of FPGA optimization techniques for circulant convolution operator and then apply quantizations to activation and element-wise operators. In the last, we propose an operator scheduling algorithm to generate the whole LSTM pipeline with the help of performance and resource models.

\subsection{Circulant Convolution Optimization}
Since the FFT based circulant convolution operator in the form of Equation~(\ref{eqn:core}) is the most computation-intensive operator in the LSTM inference model, we propose three techniques to further reduce the computational complexity by reducing the number of DFT and IDFT operator calls, and the redundant arithmetic operations of its complex number multiplication operators.

In order to reduce the number of IDFT calls in the circulant convolution operator, we propose the DFT-IDFT decoupling technique. Since DFT and IDFT are linear operators~\cite{oppenheim1999discrete}, we could decouple the DFT and IDFT operators in Equation~\ref{eqn:core} and move the IDFT operator $\mathcal{F}^{-1}(\cdot)$ outside the accumulation operator $\sum$ as following, 

\begin{equation}\label{eqn:core-decouple}
    \mathbf{a}_i=\mathcal{F}^{-1}\Big[\sum_{j=1}^q\mathcal{F}(\mathbf{w}_{ij})~\odot~ \mathcal{F}(\mathbf{x}_{j})\Big],
\end{equation}
where the number of IDFT operator calls for each circulant convolution operator is reduced from $q$ to $1$ and the numbers of the other operator calls are kept the same as before.

According to Equation~(\ref{eqn:core-decouple}), the number of calls of DFT operator $\mathcal{F}(\cdot)$ in a circulant convolution operator is $2q$, and $q$ is the number of weight vectors $\mathcal{F}(\mathbf{w}_{ij})$ and input vectors $\mathcal{F}(\mathbf{x}_j)$. Since the weight vectors $\mathbf{w}_{ij}$ are fixed when the training process is done, we could precalculate the $\mathcal{F}(\mathbf{w}_{ij})$ values and store them in the BRAM buffers of FPGAs and fetch the required values when needed instead of computing the associated DFT values at runtime. This method completely eliminates the DFT operator $\mathcal{F}(\cdot)$ calls for weight vectors and reduces the number of calls from $2qk$ to $qk$ for each circulant convolution operator. The BRAM buffer size, however, would be doubled since the outputs of DFT values $\mathcal{F}(\mathbf{w}_{ij})$ are complex numbers whose both real and imaginary parts need to be stored. In order to alleviate the BRAM buffer overhead, we propose to exploit the complex conjugate symmetry property of DFT output values, where almost half of the conjugate complex numbers could be eliminated~\cite{oppenheim1999discrete,roy2012comparison}. Therefore, there is only negligible BRAM buffer overhead to store the DFT results of weight vectors $\mathcal{F}(\mathbf{w}_{ij})$.

\begin{figure}[t!]
    \centering
    \includegraphics[width=\columnwidth]{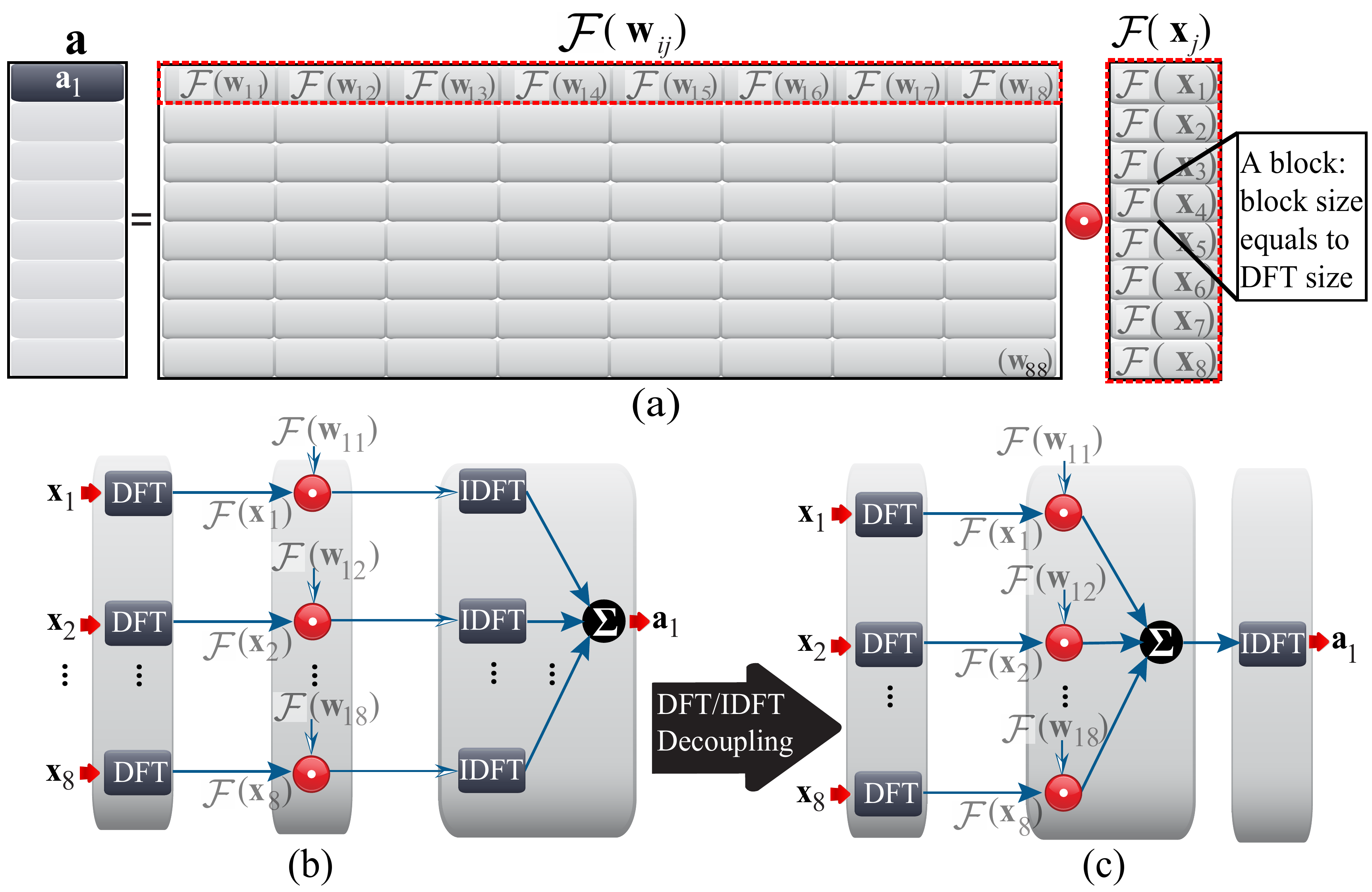}  
    \caption{An illustration of the (a) circulant convolution operator; (b) its original implementation; (c) and the optimized implementation.}
    \label{fig:Decoup}
\end{figure}

The element-wise multiplication $\odot$ between two complex number vectors $\mathcal{F}(\mathbf{w}_{ij})$ and $\mathcal{F}(\mathbf{x}_j)$ requires $4k$ multiplications and $3k$ additions. Due to the complex conjugate symmetry property of DFT $\mathcal{F}(\cdot)$ results, about half of the multiplications and additions could be eliminated. Overall, Figure~\ref{fig:Decoup} illustrates the implementations of the original and optimized circulant convolution operators when the block size is 8.

\subsection{Datapath and Activation Quantization}

The LSTM model size could be further compressed without accuracy degradation if the datapath of LSTM implementation on FPGA is carefully quantized into shorter bitwidth. We design a bit-accurate software simulator to study the impact of the bitwidth of datapath on the prediction accuracy. We first analyze the numerical range of the trained weights in the LSTM, and then determine the bitwidth of integer and fractional parts to avoid data overflow and accuracy degradation. We observe that 16-bit fixed point is accurate enough for implementing the LSTM inference model on FPGAs. 

In order to alleviate accuracy degradation problem caused by the data truncation and overflow problems in the architecture of the proposed circulant convolution operator. It is observed that the output data of IDFT are first divided by the block size (or IDFT input size) $k$, which is implemented as right shifting the numbers by $\text{log}_2 k$ bits, and then output in the last stage of IDFT pipeline. However, the more bits are right shifted, the more fractional bits are truncated and thus degrading the overall accuracy. In order to deal with the accuracy loss caused by the data truncation, we propose to evenly distribute the shift operations inside the stages of the IDFT pipeline based on the observation that right shifting one bit at a time achieves better accuracy than right shifting multiple bits at once. As for the data overflow problem, it is most likely to occur in the accumulation stage of circulant convolution operator since multiple values are summed here. We propose to move the evenly distributed right shifting operations from stages of IDFT pipeline to the ones of DFT. Since the DFT is processed before accumulation operator and right shifting makes the number to be smaller and, it is less likely to cause overflow in accumulation stage. 

\begin{figure}[t!]
    \centering
    \begin{subfigure}[t]{.49\columnwidth}
      \centering
      \includegraphics[width=\columnwidth]{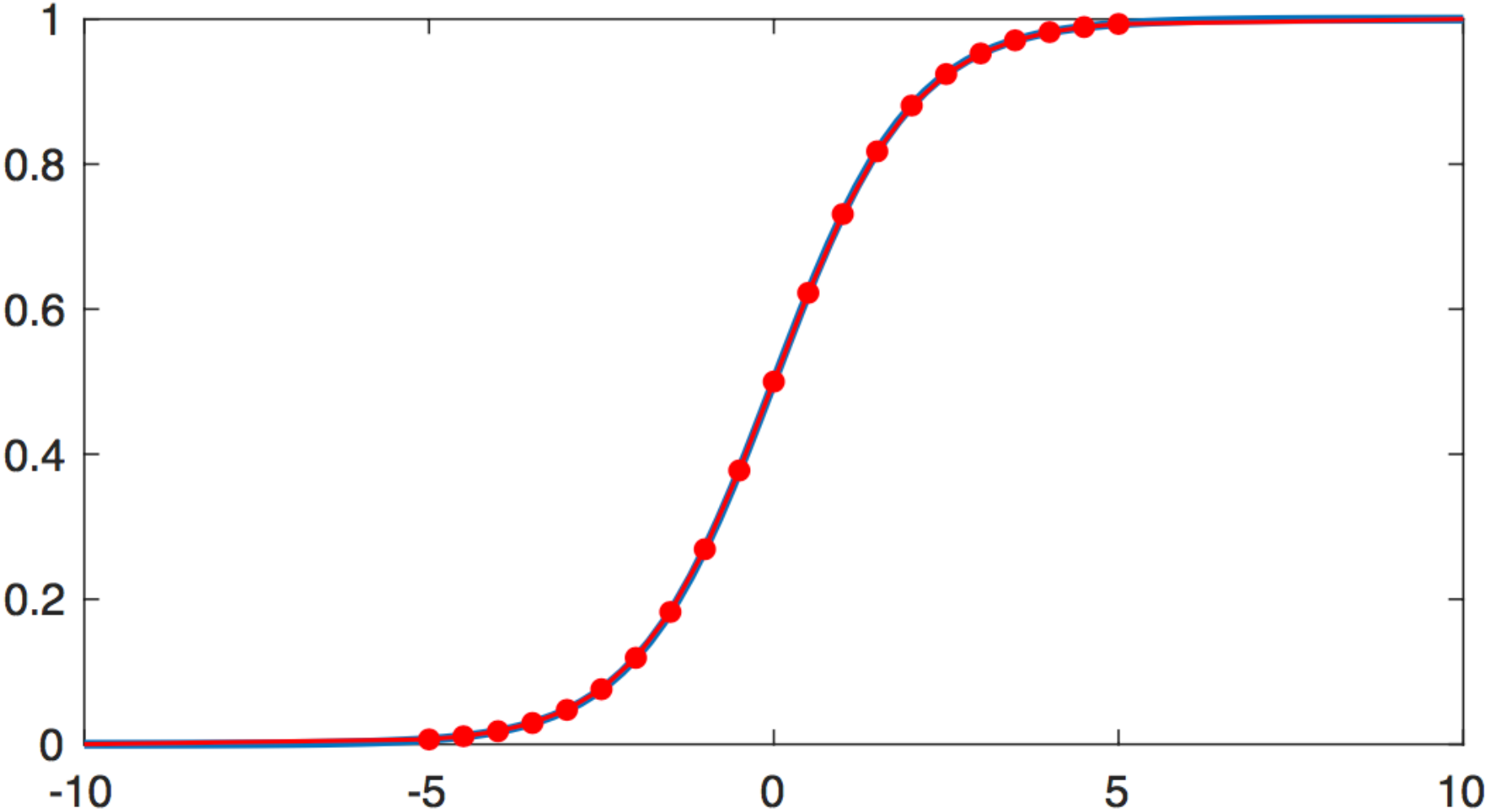}
      \caption{sigmoid}
    \end{subfigure}
    \begin{subfigure}[t]{.49\columnwidth}
      \centering
      \includegraphics[width=\columnwidth]{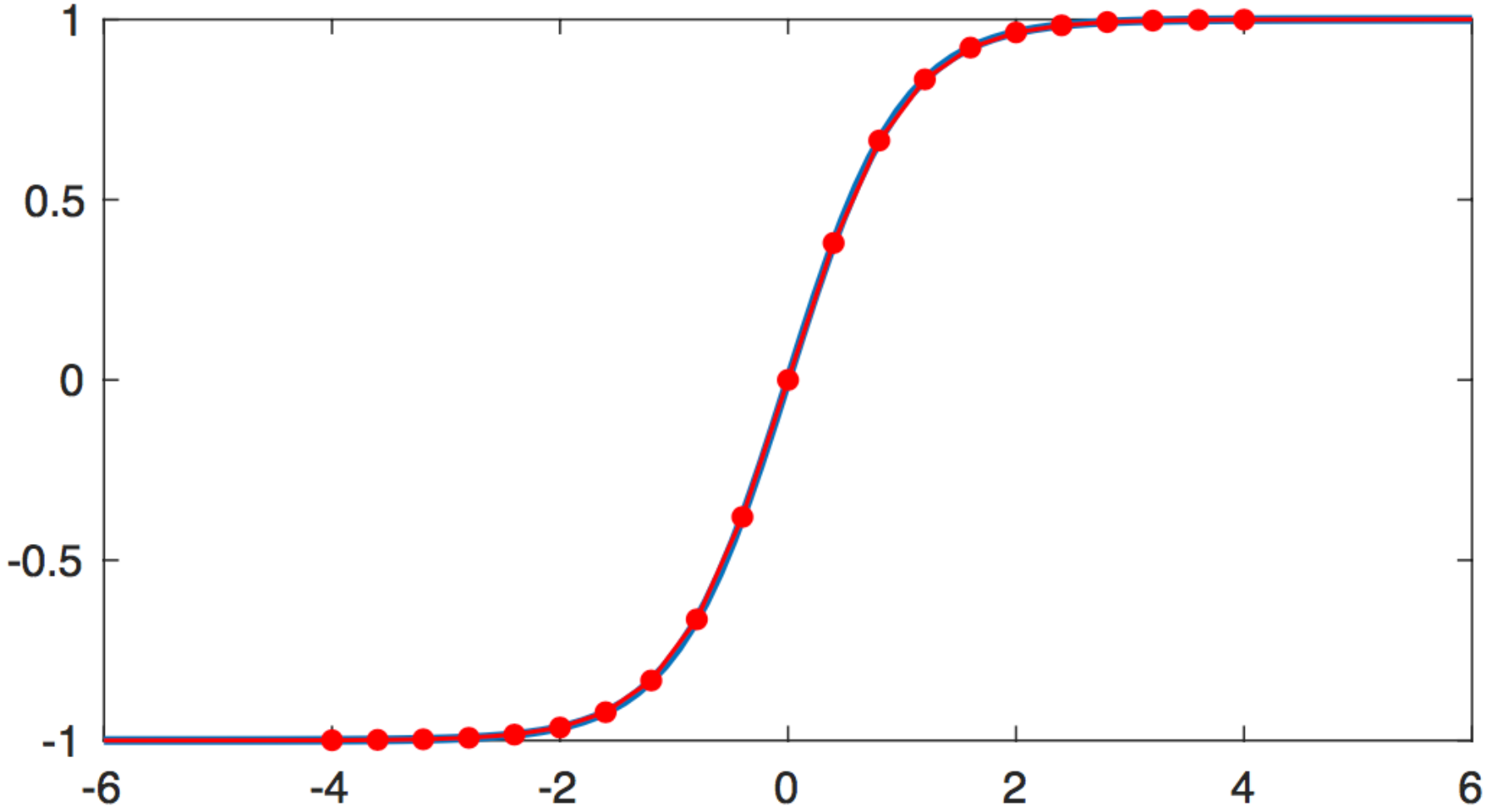}
      \caption{tanh}
    \end{subfigure}
    \caption{Piece-wise linear activation functions.}
    \label{fig:activation_function}
 \end{figure}

The activation functions in LSTMs are all transcendental functions whose implementations on FPGA are very expensive with respect to resource utilization. In order to achieve a balance between accuracy and resource cost, we propose to utilize quantized piece-wise linear functions to approximate them. Figure~\ref{fig:activation_function} shows that the sigmoid and tanh functions are approximated using piece-wise linear functions with 22 segments. As we can see from the figure, the approximated and the original functions are almost the same and the error rate is less than 1\%. Since the linear function could be represented in the slope-intercept form like $y = ax+b$, we only need to store the associated slope $a$ and intercept $b$ for each piece of linear function. In the real implementation, the computational complexity of activation functions only involves a simple comparison to index the associated pair of slope and intercept and one 16-bit fixed point multiplication followed by an addition. It is necessary to note that, according to our experimental results, the piece-wise linear approximation incurs negligible accuracy degradation for LSTMs. 

\subsection{Operator Scheduling}

\begin{figure*}[t!]
    \centering
    \includegraphics[width=1.8\columnwidth]{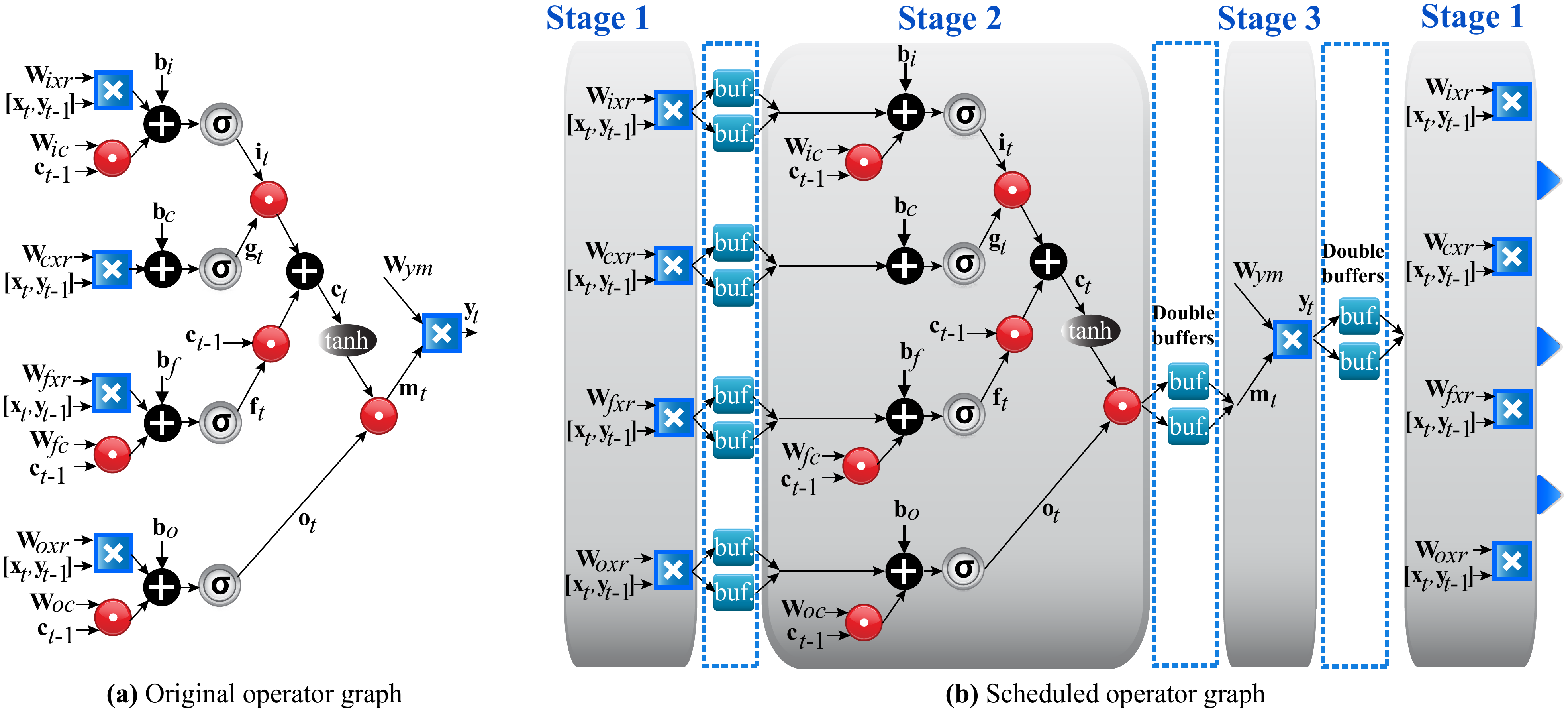} 
    \caption{Illustration of operator scheduling on data dependency graph. The circle represents the element-wise operator, and the square represents the circulant convolution operator.}
    \label{fig:op_fusion_scheduling}
\end{figure*}

The recurrent nature of LSTM enforces strict data dependency among operators inside the LSTM module. In order to accommodate the complicated interactions of LSTM primitive operators, we propose a graph generator to transform the LSTM algorithm specification in the form of the equations like Equation~(\ref{eqn:model}) to a directed acyclic data dependency graph. Figure~\ref{fig:op_fusion_scheduling} (a) shows the generated LSTM directed operator graph from the LSTM descriptions, where each node is an LSTM primitive operator and the edge represents the data dependency between two operators. It is necessary to note that the generated graph is acyclic because we deliberately remove the feedback edges from cell output $\mathbf{c}_t$ to the LSTM module output $\mathbf{y}_t$. Since the backward edges are taken care of by the double-buffer mechanism, this practice would never harm the correctness and efficiency of the final LSTM accelerator design.

\begin{figure}[t!]
    \centering
    \includegraphics[width=.62\columnwidth]{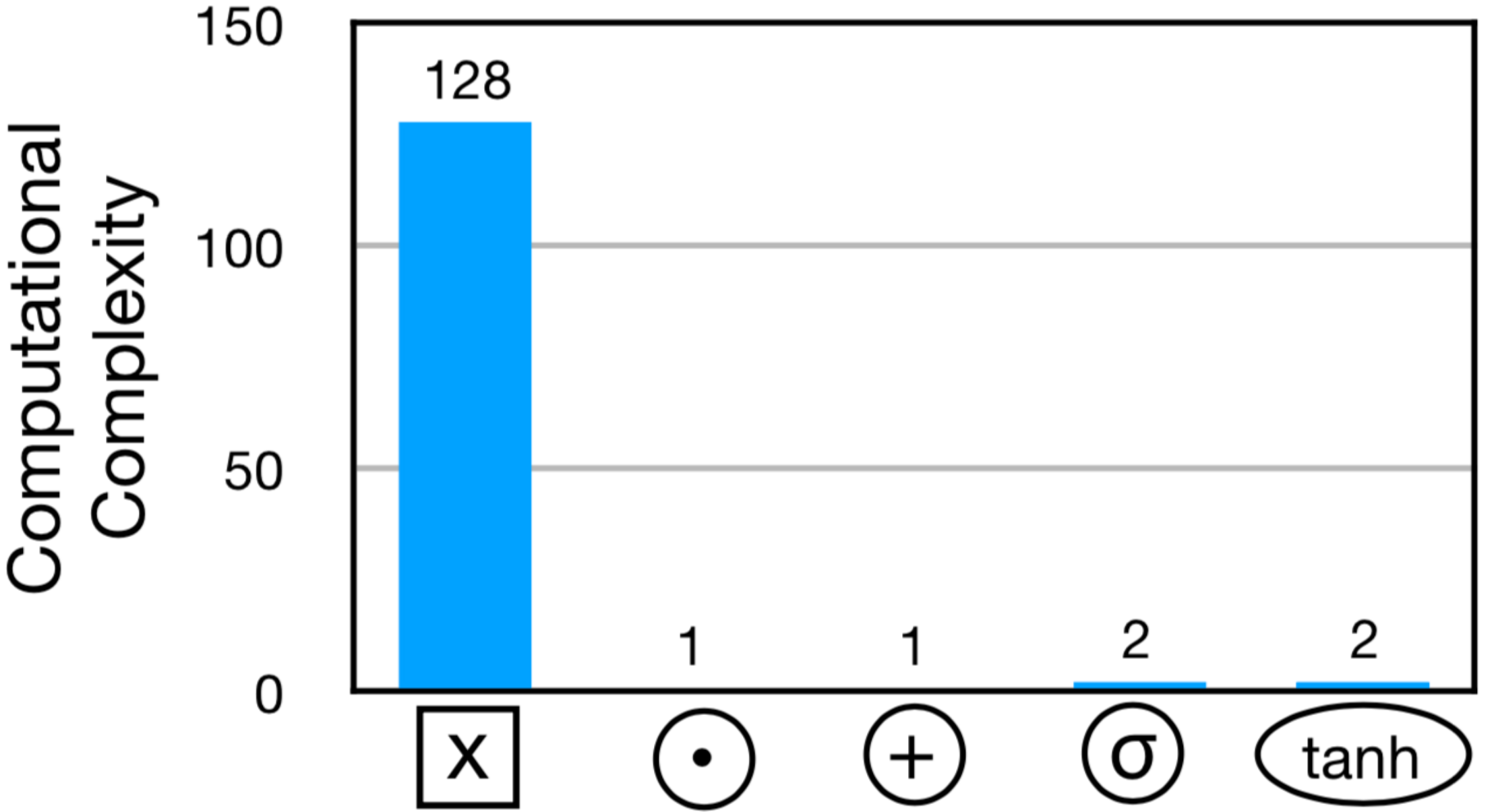} 
    \caption{Computational complexity of LSTM operators.}
    \label{fig:op_compute_complexity}
 \end{figure}

LSTMs exhibit a highly skewed distribution of computation complexity among the primitive operators. Figure~\ref{fig:op_compute_complexity} shows the normalized computational complexity of the five primitive operators of the Google LSTM~\cite{sak2014long} studied in this work. The computational complexity gap between the circulant convolution operator and element-wise multiply operator $\odot$ is as large as 128 times. So, if we want to pipeline these two operators we must either boost the parallelism of the former operator or make the latter operator wait (idle) for the former one. However, the reality is that the limited on-chip resources of FPGAs generally cannot sustain sufficient parallelism and the idle operators make the design inefficient. Therefore, pipelining a complex LSTM algorithm as a whole, such as the Google LSTM~\cite{sak2014long} shown in~\ref{fig:op_fusion_scheduling}(a), is very inefficient on FPGAs.

 In order to deal with this problem, we propose to break down the original single pipeline into several smaller coarse-grained pipelines and overlap their execution time by inserting double-buffers for each concatenated pipeline pair. For example, the original operator graph of Google LSTM~\cite{sak2014long} in~\ref{fig:op_fusion_scheduling}(a) is divided into three stages in~\ref{fig:op_fusion_scheduling}(b), where each stage will be implemented as a coarse-grained pipeline on FPGAs. The double-buffers added between stages are used to buffer the data produced/consumed by the previous/current stage. However, scheduling the operators to different stages in an efficient way is still a problem. We propose an operator scheduling algorithm shown in Algorithm~\ref{algorithm:scheduling} to tackle this problem. The algorithm takes the original operator graph $G=(V,E)$, operator weight set $W(V)$, and operator priority set $P(V)$ as input and outputs several operator subgraphs $G_k$. For original operator graph $G=(V,E)$, each vertex $v_i\in V$ represents an operator and the edge $e_{ij}$ represents the data dependency between $v_i$ and $v_j$. Each vertex $v_i$ has a weight $W(w_i)$ which is the associated arithmetic computational complexity. The algorithm first traverses down the graph from the source vertex computing the priority of each vertex by 

 \begin{equation} \label{eqn:priority}
 P(v_i) = \begin{cases} W(v_i) + \max\limits_{v_j\in Succ(v_i)}{P(v_j)},\ \ v_i \neq v_{sink} \\
 W(v_{sink}), \ \ \ \ \ \ \ \ \ \ \ \ \ \ \ \ \ \ \ \ \ \ \ \text{otherwise}
\end{cases}
 \end{equation} 
 Since $P(v_i)$ is accumulated with the maximum value of successors $P(v_j)$ as shown in Equation~(\ref{eqn:priority}), priority set $P(V)$ is topologically ordered, which means that it is guaranteed that all predecessor operators are scheduled before scheduling a new operator~\cite{lee2015orchestrating}. After the prioritization, the algorithm selects the operator with the highest priority value and then determines the parallelism of the operator $N(v_j)$ and whether it should be added to the current or a new stage according to the resource utilization of FPGAs. Then, the operator subgraphs $G_k$ and the operator parallelism set $N(V)$ are output by this algorithm, where each stage represents a corresponding LSTM execution stage that will be implemented as a coarse-grained pipeline on FPGAs. Since the overall throughput of this coarse-grained pipeline design is constrained by the slowest stage, we need to further determine the pipeline replication factor $R(G_k)$ for each stage. To fully utilize the resources of a certain FPGA chip, we also need to take into account of the resource utilization of each stage, and thus we propose to enumerate pipeline replication factor $R(G_k)$ to get the optimal setting with the help of our analytical performance and resource models which are presented in Section~\ref{sec:models}.

\begin{algorithm}[!tb]
\scriptsize
 \KwIn{operator graph $G = (V, E)$, operator weight set $W(V)$, and priority set $P(V)$;}
 \KwOut{operator subgraph of each stage $G_k = (V_k, E_k)$;}
 Traverse $G=(V,E)$ and compute priority set $P(V)$;\\
 $k \leftarrow 0, N(V) \leftarrow \{\textbf{1}\}$; \\
 \ForEach{$v_i \in$ V in decreasing order of $P(v)$}{
    \uIf{$k = 0$}{
        $k \leftarrow k + 1$;\\
        $G_k \leftarrow v$; \tcp{add the operator to a new stage}
    }
    \Else{
        \ForEach{$N'(v_j) \in G_k$}{
            $N'(v_j) \leftarrow N(v_j)\cdot \ceil{\frac{W(v_j)}{W(v_i)}}$;
        }
    \uIf{resource constraints are satisfied} {
    $G_j \leftarrow v$; \tcp{add the operator to current stage}
    $N(V) \leftarrow N'(V)$; \tcp{update operator parallelisms}
    } 
    \Else{$k \leftarrow k+1$; \\
    $G_k \leftarrow v_i$; \tcp{add the operator to a new stage}}
    }
 }
 $K \leftarrow k$; \\
 Enumerate $R(G_k)$ values to maximize throughput and fully utilize FPGA resource;\\
 \Return{$N(V)$, \{$G_1, G_2, ..., G_K$\}, and \{$R(G_1), R(G_2), ..., (G_K)$\}};\\
 \caption{Operator Scheduling Algorithm}\label{algorithm:scheduling}
 \end{algorithm}

\subsection{Performance and Resource Models}\label{sec:models}

 Since the throughput of the proposed coarse-grained pipeline design is constrained by the slowest stage, the analytical performance model is built as following,  

 \begin{equation}
 FPS = \frac{Frequency}{\max{\{T_1, T_2,...,T_h,...T_K\}}},
 \end{equation}
where $FPS$ is the number of frames per second of C-LSTM accelerator, $T_k$ represents the number of execution clock cycles of stage $k$, and $K$ is the total number of stages. $T_k$ is calculated by considering the parallelism and input data size of each stage as following,


\begin{equation}
T_k = \ceil{\max\limits_{v_i \in G_k}{\frac{Q(v_i)}{N(v_i)}}/R(G_k)} + D_k
\end{equation}
where $Q(v_i)$ is the workload of operator $v_i$ and $D_k$ is the pipeline depth of stage $k$. It is necessary to note that, the compression ratio of the block-circulant matrices based technique is large enough to store the whole LSTM model on BRAMs of FPGAs, and for each frame, we only need to load very limited size of input data which makes computation time of LSTM to be overlapped with data loading.

The resource model of the highly optimized primitive operator templates is very straightforward because the linear model with respect to the associated operator parallelism $N(v_i)$ and stage parallelism $R(G_k)$ is accurate enough to guide the design space exploration for energy-efficient designs. The models are shown in the following,

\begin{gather}
DSP = \sum_{k=1}^{K} R(G_k) \cdot \sum_{v_i \in V,}\Delta DSP(v_i) \cdot N(v_i),\\
BRAM = \sum_{k=1}^{K} R(G_k) \cdot \sum_{v_i \in V,}\Delta BRAM(v_i) \cdot N(v_i),\\
LUT = \sum_{k=1}^{K} R(G_k) \cdot \sum_{v_i \in V,}\Delta LUT(v_i) \cdot N(v_i),
\end{gather}
where $\Delta DSP(v_i)$, $\Delta BRAM(v_i)$, and $\Delta LUT(v_i)$ are obtained by profiling the resource consumption values for operator $v_i$ on the FPGA using the manually optimized operator template.

\subsection{Putting It All Together}

\begin{figure}[t!]
    \centering
    \includegraphics[width=.9\columnwidth]{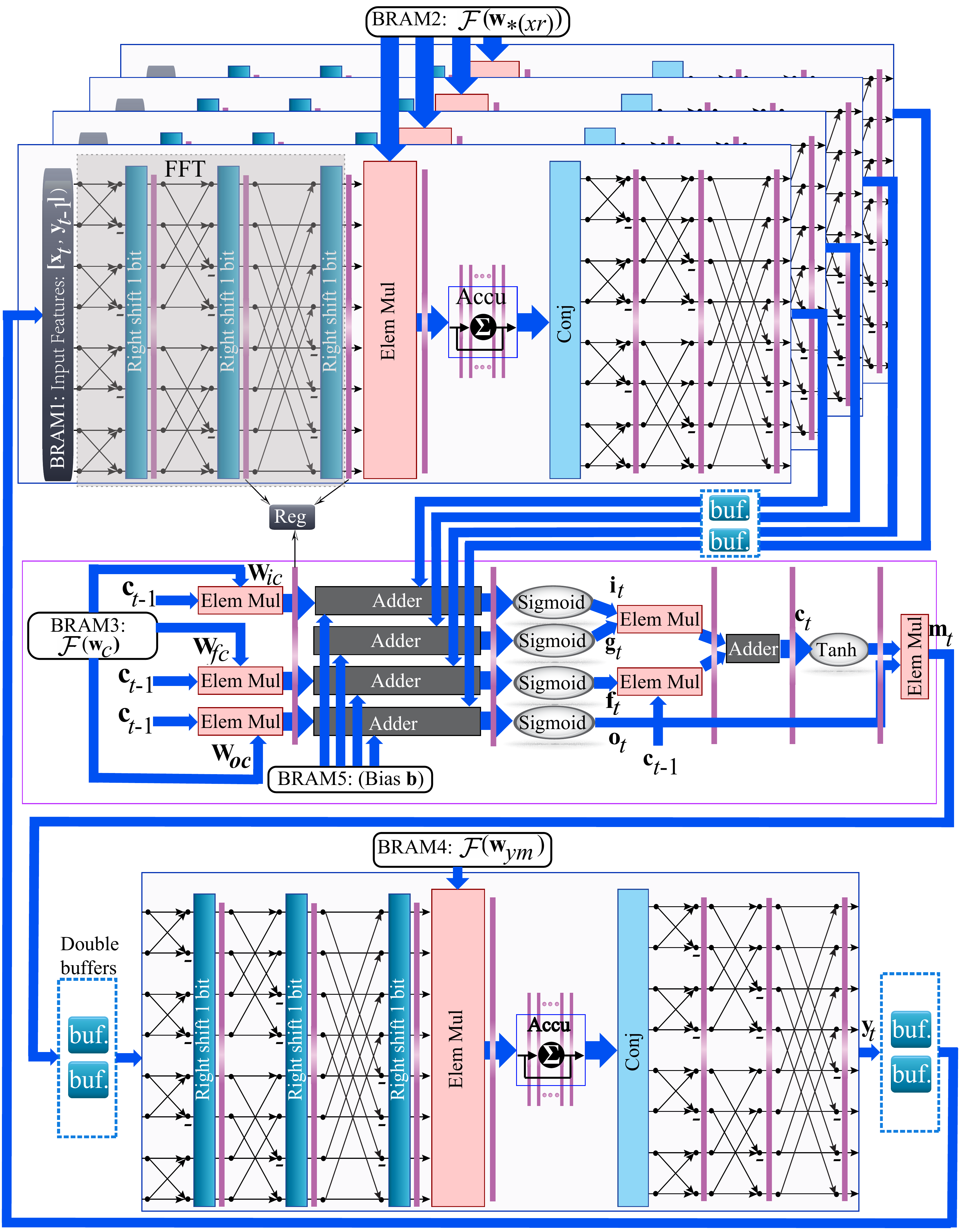} 
    \caption{The proposed Google LSTM architecture.}
    \label{fig:arch}
\end{figure}

The final hardware architecture of the Google LSTM algorithm~\cite{sak2014long} is shown in Figure~\ref{fig:arch}. This design mainly consists of three coarse-grained pipeline stages corresponding to the operator scheduling result shown in Figure~\ref{fig:op_fusion_scheduling}(b). At Stage 1, the input vectors $\mathbf{x}_t$ and the prestored DFT values of weight  matrices $\mathbf{W}$ are convolved using the circulant convolution operator whose output is written into the double-buffer. Since all the DFT values of weight matrices are compressed small enough, they could be stored in on-chip BRAM buffers instead of off-chip DDR memory. The performance of the circulant convolution operator is thus no longer bottlenecked by off-chip memory bandwidth and the parallel compute units could be fully exploited on FPGAs. In stage 2, the input data are first read from double-buffer of the previous stage and then processed by a series of element-wise operators including addition, multiplication and activation functions in the LSTM cell module. The output of Stage 2 is also written to double-buffer for the next stage. As for Stage 3, the results of the prior stage are fetched from double-buffer and are then projected to output using the circulant convolution operator. In the last, the projected output will be forwarded to Stage 1 for the next iteration.


\begin{figure*}[t!]
    \centering
    \includegraphics[width=1.8\columnwidth]{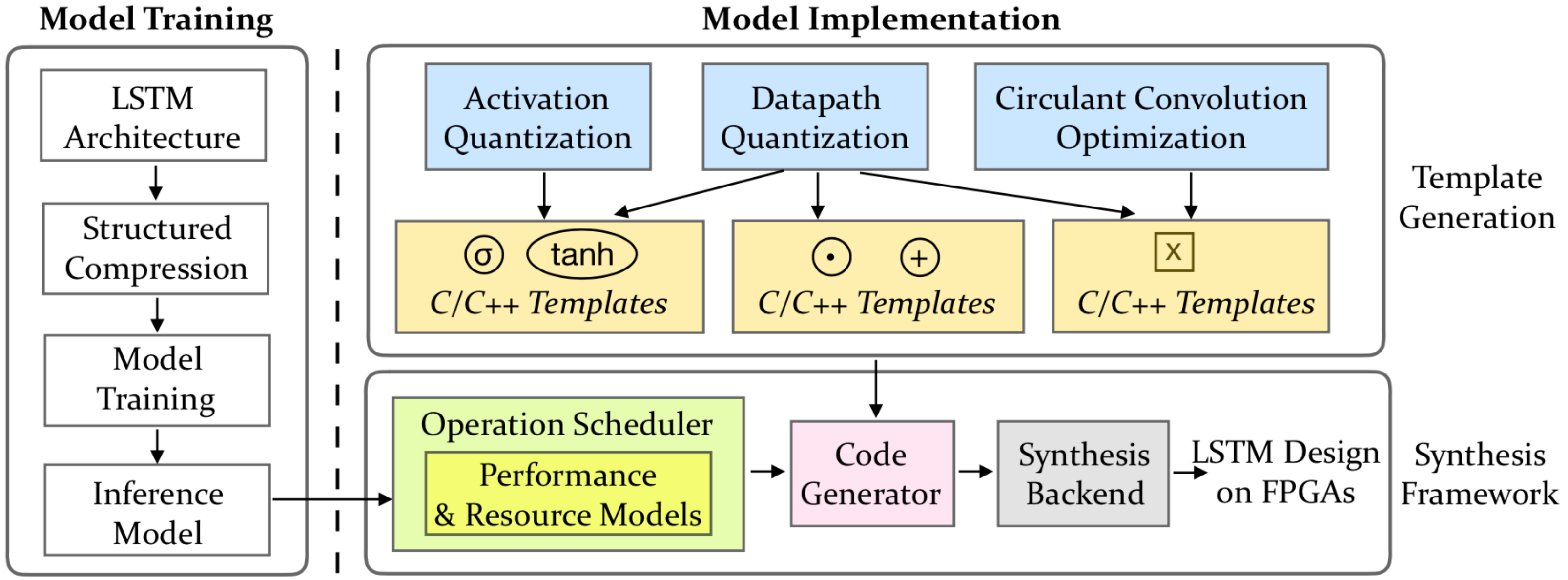} 
    \caption{C-LSTM framework overview.}
    \label{fig:opt_framework}
 \end{figure*}

\section{C-LSTM Framework}
In order to embrace a wide range of LSTM architectures, we propose a comprehensive framework called C-LSTM to assist the LSTM model training using the block-circulant matrix based structured compression and enable an automatic flow to generate efficient LSTM inference designs on FPGAs. As shown in Figure~\ref{fig:opt_framework}, the C-LSTM framework is mainly composed of two parts which are LSTM model training and its implementation on FPGAs. The details of the C-LSTM framework are explained in the following sections.

\subsection{Model Training}
The model training, which is shown on the left side of Figure~\ref{fig:opt_framework}, accepts the LSTM architecture specifications in the form of Equation~(\ref{eqn:model}) as input. Then, the block-circulant matrix based structured compression is applied to the weight matrices of the model. In the following, TensorFlow~\cite{abadi2016tensorflow} is used as the training framework to iteratively train the LSTM model. The trade-offs between compression ratio and prediction accuracy is explored in this procedure. In the last, the LSTM inference model is configured with the well-trained weight matrices and sent to model implementation flow for further acceleration on FPGAs.

\subsection{Model Implementation}

The model implementation of the C-LSTM framework is shown in the right side of Figure~\ref{fig:opt_framework}, It mainly consists of two parts which are operator templates generation (upper part) and automatic synthesis framework (lower part). Since the number of primitive operators of LSTMs is limited, we propose to manually write the template for each primitive operator. As for the LSTM algorithms studied in this work, we define hyperbolic tangent $tanh$, sigmoid $\sigma$, element-wise vector addition, element-wise vector multiplication, and circulant convolution as primitive operators. The optimization techniques presented in Section~\ref{sec:hardware_acc} are all applied to these operators. It is necessary to note that, the proposed primitive operator templates are general enough to implement almost any kind of LSTM variant to the best of our knowledge.

The automatic synthesis framework is fed with the well-trained inference model provided by the model training flow. Then a directed acyclic data dependency graph is generated to represent the computation flow of LSTM. The operators in the graph are scheduled to compose a multi-stage coarse-grained pipeline as to maximize the performance under certain resource constraints with the help of analytical performance and resource models. The scheduling result is then given to the code generator. The code generator takes the operator scheduling result as input and generates the final C/C++ based code automatically by integrating the associated primitive operator templates together. Since the interface of each template is well defined and the tunable parameters are expressed using C/C++ marcos, the code generation is very efficient. The synthesis backend which is an off-the-shelf commercial HLS tool, accepts the C/C++ code as input and outputs the optimized LSTM hardware implementation on FPGAs. It is necessary to note that each commercial HLS toolchain requires specific coding style to achieve the best performance, and thus the templates of the primitive operators should be tailored accordingly~\cite{CodingSytle:2017:FPGA}.

\begin{table}[!t]
    \centering
    \caption{Comparison of FPGA platforms}\label{tbl:platform}
    \resizebox{\columnwidth}{!}{
        \begin{tabular}{|c|c|c|c|c|c|}
            \hline
            FPGA & DSP & BRAM & LUT & FF & Process \\ 
            \hline

            XCKU060 & 2,760 & 1,080 & 331,680 & 663,360 & 20nm\\
            \hline

            Virtex-7(690t) & 3,600 & 1,470 & 859,200 & 429,600 & 28nm\\
            \hline        
        \end{tabular}
    }
\end{table}
 
\begin{table*}[t]
  \centering
  \caption{Detailed comparison for different LSTM designs.} 
  \label{table:Exp-Result}
  \begin{threeparttable}
  \begin{tabular}{|c|c|c|c|c|c|c|c|c|c|}
    \hline
   \  & \textbf{ESE~\cite{han2017ese}} & \multicolumn{2}{c|}{\makecell{\textbf{C-LSTM FFT8}\\(Block size: 8)}} & \multicolumn{2}{c|}{\makecell{\textbf{C-LSTM FFT16}\\(Block size: 16)}} &  \multicolumn{2}{c|}{\makecell{\textbf{C-LSTM FFT8}\\(Block size: 8)}} & \multicolumn{2}{c|}{\makecell{\textbf{C-LSTM FFT16}\\(Block size: 16)}} \\
   \hline
   \textbf{\makecell{LSTM \\Algorithm}} & \multicolumn{5}{c|} {Google LSTM~\cite{sak2014long}} & \multicolumn{4}{c|} {Small LSTM~\cite{small_lstm}}\\
   \hline

   \textbf{\makecell{Weight Matrix Size\\(\#Parameters of LSTM)}} & 0.73M & \multicolumn{2}{c|}{0.41M}& \multicolumn{2}{c|}{0.20M} & \multicolumn{2}{c|}{0.28M} & \multicolumn{2}{c|}{0.14M} \\
   \hline
  
   \textbf{Quantization} & 12bit fixed & \multicolumn{2}{c|}{16bit fixed} & \multicolumn{2}{c|}{16bit fixed} & \multicolumn{2}{c|}{16bit fixed} & \multicolumn{2}{c|}{16bit fixed}\\ 
   \hline

   \textbf{\makecell{Matrix \\ Compression Ratio}} & 4.5 : 1\tnote{1} & \multicolumn{2}{c|}{7.9 : 1} & \multicolumn{2}{c|}{15.9 : 1} &  \multicolumn{2}{c|}{7.9 : 1} & \multicolumn{2}{c|}{15.9 : 1}\\
   \hline

   \textbf{Platform} & KU060 & KU060 & 7V3 & KU060 & 7V3 & KU060 & 7V3 & KU060 & 7V3\\
   \hline

   \textbf{DSP (\%)}  & 54.5 & 96.5 & 74.3 & 98.0 & 77.4 & 77.6 & 60.5 & 84.9 & 65.2\\
   \hline

   \textbf{BRAM (\%)} & 87.7 & 87.6 & 65.7 & 89.1 & 63.3 & 83.3 & 66.9 & 87.2 & 64.1 \\
   \hline

   \textbf{LUT (\%)}  & 88.6 & 75.2 & 58.7 & 72.8 & 55.3 & 92.5 & 67.6 & 93.6 & 72.3\\
   \hline

   \textbf{FF (\%)}   & 68.3 & 58.9 & 46.5 & 63.4 & 48.1 & 61.2 & 49.0 & 70.7 & 54.6\\
   \hline

   \textbf{Frequency (MHz)} & \multicolumn{9}{c|}{200}\\
   \hline


   \textbf{\makecell{PER Degradation}} & 0.30\% & \multicolumn{2}{c|}{0.32\%} & \multicolumn{2}{c|}{1.23\%} & \multicolumn{2}{c|}{0.29\%} & \multicolumn{2}{c|}{1.16\%}\\
   \hline

   \textbf{Latency ($\mu$s)} & 57.0 & 15.4 & 16.7 & 8.1 & 9.1 & 8.9 & 9.8 & 4.8 & 5.4 \\
   \hline

   \textbf{\makecell{Frames per \\Second (FPS)}} & 17,544 & 195,313 & 179,687 & 371,095 & 330,275 & 337,838 & 307,432 & 628,379 & 559,257 \\
   \hline

   \textbf{Power (W)} & 41 & - & 22 & - & 23 & - & 21 & - & 22\\
   \hline

   \textbf{\makecell{Energy \\Efficiency \\(FPS/W)}} & 428 & - & 8,168 & - & 14,359 & - & 14,640 & - & 25,420\\
   \hline

  \end{tabular}
\begin{tablenotes}
\item[1] This estimation considers both weights and indices (there is at least one index per weight after compression in ESE). \\However, this is a pessimistic estimation for ESE because indices can use fewer bits for representation than weights;
\end{tablenotes}
\end{threeparttable}
\end{table*}

\section{Experiment Evaluation}

\subsection{Experiment Setup}

The proposed techniques for LSTMs are evaluated on two platforms: Xilinx KU060 and Alpha Data's ADM-7V3. The Xilinx KU060 platform consists of a Xilinx XCKU060 FPGA and two 4GB DDR3 memory. The ADM-7V3 board consists of a Xilinx Virtex-7 (690t) FPGA and a 16GB DDR3 memory. The comparison of the FPGA on-chip resources of the two platforms is presented in Table~\ref{tbl:platform}. The ADM-7V3 FPGA board is connected to the host via PCI-e 3.0 X8 interface, and the host machine is a server with Intel Core i7-4790 CPU. Xilinx SDx 2017.1 is used as the commercial synthesis backend to synthesize the C/C++ based LSTM design onto FPGAs. The proposed FPGA implementations of LSTMs are operating at 200MHz on both platforms.

We measure the latency of our C-LSTM designs on KU060 platform using the number of clock cycles times the clock period (5ns) reported by Xilinx SDx tools. To make a fair comparison with ESE~\cite{han2017ese}, the latency of ESE reported in Table~\ref{table:Exp-Result} is its theoretical time. Since we do not have the KU060 platform, we cannot give out an accurate estimation and the associated power and energy efficiency results are left blank. As for the ADM-7V3 platform, the execution time of C-LSTM designs are obtained by using Xilinx SDx runtime profiler, and the power is profiled using the TI Fusion Power device through the associated interface on ADM-7V3 with a sampling rate of 100Hz. 

Besides the LSTM based RNN architecture used in \cite{han2017ese,sak2014long}, we also evaluated the performance on a smaller LSTM model~\cite{small_lstm}, where the input feature is a $39$-dimension vector ($12$ filterbank coefficients plus energy and its first/second temporal derivatives), and the gate/cell layers' dimension is $512$. In this small model, the peephole connection and projection layer are not employed. The model contains two stacked LSTM as well. However, we used bidirectional architecture~\cite{yu2014automatic,graves2013hybrid} to get a better PER.

In order to make a convincing validation for the superiority of the proposed C-LSTM optimization framework, we compare our design with the state-of-the-art LSTM design ESE~\cite{han2017ese}.The same dataset, LSTM algorithm, and FPGA platforms are used in the associated experiments as ESE to make a fair comparison. 

\subsection{Experimental Results of Google LSTM}

With the compression technique of C-LSTM, we are able to store all the weights matrices and the projection matrix in BRAM, after performing compression on the baseline. The baseline has the same structure as the baseline in ESE.


According to the results of latency and FPS in Table~\ref{table:Exp-Result}, we achieve 3.6X and 4.3X latency reduction and 11X and 13X performance speedup for FFT8 and FFT16 based compression techniques compared with ESE on the platform of KU060. It is necessary to note that the gap between latency reduction and performance speedup stems from the coarse-grained architecture of the proposed LSTM accelerator. And thus the latency of our proposed C-LSTM accelerator for Google LSTM algorithm is the latency of one stage multiplied by 3, because each input frame needs to go through three coarse-grained pipelines. However, after three frames have been processed, the following frame could be processed at every one stage of latency.


As we can see from Table~\ref{tbl:platform}, the resource of the FPGA chip Virtex-7 of the ADM-7V3 platform is 30\% higher than the FPGA XCKU060 of KU060 platform. Therefore, to make a fair comparison, we use the total resource of KU060 as the resource consumption bound for the ADM-7v3 platform. Compared with ESE, we achieve 10.2X and 18.8X performance speedups and 19.1X and 33.5X energy efficiency gains using FFT8 and FFT16, respectively. Since the power consumption of C-LSTM is only half of the ESE, the energy efficiency gain is higher than performance. It is necessary to note that as shown in Table~\ref{tbl:platform}, the manufacturing process of XCKU060 FPGA is 20nm while the process of Virtex-7 is 28nm, which means the energy efficiency gain reported here is pessimistic.

Although the promising performance and energy gains are achieved by C-LSTM, the resource utilization for LUT, FF, and BRAM are less than ESE, and more important, the relative PER degradation is very small, which are 0.32\% and 1.23\% using FFT8 and FFT16, respectively. After detailed analysis, we summarize the fundamental reasons for the high performance and power gains in three aspects. First, the structured compression used in this work eliminates the irregular computation and memory accesses which not only makes the design more regular but also exposes more parallelism. This could be verified in that the DSP resource consumption of the proposed method is much more than ESE. Secondly, the whole model (weights matrices and the projection matrix) could be stored on-chip without fetching data from off-chip DRAM, making the LSTM not bounded by memory. Lastly, the more efficient implementation of LSTM on FPGAs contributes to the high efficiency. For example, we use the 22-segment piece-wise linear function to approximate the activation functions while ESE employs look-up tables which break the activation down into 2048 segments and consume more resources. Moreover, we propose to employ FFT based block-circulant matrix multiplication while ESE uses sparse matrix multiplication which needs to store extra indices for sparse matrices and thus prevents from storing the whole model on-chip.

\subsection{Experimental Results of Small LSTM}

In order to validate that proposed C-LSTM is not only appropriate for Google LSTM model, we also implement a Small LSTM~\cite{small_lstm} model on both FPGA platforms.

In KU060 platform, the FFT8 and FFT16 designs could achieve 19.3X and 35.9X performance speedup compared with ESE, respectively. In the ADM-7V3 platform, the performance speedups are 17.5X and 31.9X and the energy efficiency gains are 34.2X and 59.4X compared with ESE, respectively. For both platforms, the PER degradation is 0.29\% and 1.16\% for FFT8 and FFT16, respectively.

\section{Related Work}
Recently, FPGA has emerged as a promising hardware acceleration platform for DNNs as it provides high performance, low power and reconfigurability. A lot of FPGA based accelerators have been proposed for convolutional neural networks (CNNs) to overcome the computing and energy efficiency challenges. \cite{Wei:2017:DAC} proposes to utilize systolic array based convolution architecture to achieve better frequency and thus performance for CNNs on FPGAs. \cite{lu2017evaluating} employs the Winograd algorithm to reduce the multiplication operators as to save DSP resources and accelerate matrix multiplication in CNNs. \cite{Xiao:2017:DAC} proposes to take advantage of the heterogeneous algorithms to maximize the resource utilization for convolutional layers on FPGAs. Some studies also propose to transform the CNN models to frequency domains and then exploit FFT algorithms for further acceleration~\cite{Ko:2017:DAC}. The FFT based acceleration scheme used in the CNN model is completely different from this work, in which we target on a totally different LSTM based RNN model and the FFT algorithm is applied to the circulant convolution operators instead of the convolution layers of CNNs. 

There are also a lot of works on implementing RNN accelerators for FPGAs~\cite{guan2017fpga,li2015fpga,nurvitadhi2016accelerating}. \cite{nurvitadhi2016accelerating} designs an accelerator for the gated recurrent network (GRU) which embodies a different architecture from the LSTM based RNNs. \cite{guan2017fpga} and \cite{li2015fpga} focus on LSTM based RNNs but none of these works utilize compression techniques to reduce the model size. The most relevant study to this work is ESE~\cite{han2017ese}, which proposes a software and hardware co-design framework to accelerate compressed sparse LSTM model obtained by parameter pruning~\cite{han2015deep}. The performance and energy efficiency gains achieved by ESE is very promising compared with CPU and GPU based implementations. However, due to the irregular computation and memory accesses caused by the sparse weight matrices of the compressed model, the computing power of the FPGA is not fully exerted by ESE. In order to deal with this problem, this work proposes to employ a structured compression technique as to completely eliminate the irregularities of computation and memory accesses. Moreover, a suite of highly efficient optimization techniques is enabled by an automatic synthesis framework to generate LSTM accelerators with much higher performance and energy efficiency under the same conditions.

\section{Conclusion}
In this paper, we propose to employ a structured compression technique using block-circulant matrices to compress the LSTM model small enough to be fitted on BRAMs of FPGA. Besides the reduced model size, the irregular computation and memory accesses have been completely eliminated by the regular structure of the block-circulant matrices. Moreover, an efficient FFT based fast circulant convolution is applied to accelerate the LSTM computation by reducing both the computational and storage complexities. In order to accommodate a wide range of LSTM variants, we also propose an automatic optimization and synthesis framework. Overall, compared with the state-of-the-art LSTM implementation, the proposed C-LSTM designs generated by our framework achieve up to 18.8X and 33.5X gains for performance and energy efficiency with small accuracy degradation, respectively.

\begin{acks}
This work is supported by Beijing Natural Science Foundation (No. L172004) and National Science Foundation under grants CNS \#1704662 and CNS \#1739748. We thank all the anonymous reviewers for their feedback.
\end{acks}

\printbibliography

\end{document}